%% file: root.tex
\documentclass[letterpaper, 10 pt, conference]{ieeeconf}  % Comment this line out if you need a4paper

\IEEEoverridecommandlockouts                              % This command is only needed if 
\usepackage{amsfonts}
\usepackage{amsmath, float}
\usepackage{graphicx}

\usepackage{colortbl}
\usepackage{booktabs}
\graphicspath{{Figures/}}
\usepackage{amssymb,bbm, array}
\usepackage[table,xcdraw]{xcolor}
\usepackage{multirow}
\usepackage{adjustbox}
\usepackage{multirow}
\usepackage{bbding, pifont}
\usepackage[hang,flushmargin]{footmisc}
\usepackage[breaklinks,colorlinks]{hyperref}

\definecolor{darkred}{rgb}{0.7, 0, 0}
\definecolor{darkblue}{rgb}{0.0, 0.0, 0.7}
\definecolor{darkgreen}{rgb}{0, 0.4, 0}

\usepackage{cite}
\let\labelindent\relax
\usepackage{enumitem}
\usepackage{tabularray}
\usepackage[hang,flushmargin]{footmisc}
\usepackage[breaklinks,colorlinks]{hyperref}

\UseTblrLibrary{siunitx}

\usepackage{color}
\usepackage{xcolor}
\usepackage{caption}
\newcommand{\modelname}{\textsc{Draw2Act}}
\usepackage[misc]{ifsym}

\title{\LARGE \bf
\modelname: Turning Depth-Encoded Trajectories into Robotic Demonstration Videos
}

\author{Yang Bai$^{1,2}$, Liudi Yang$^{3}$, George Eskandar$^{5}$, Fengyi Shen$^{4}$, \\
Mohammad Altillawi$^{5}$, Ziyuan Liu$^{5\textrm{ \Letter}}$\thanks{\textrm{\Letter} : Corresponding author.}, Gitta Kutyniok$^{1,2}$ \\
$^{1}$Ludwig Maximilian University of Munich, $^{2}$Munich Center for Machine Learning (MCML)\\$^{3}$University of Freiburg, $^{4}$Technical University of Munich, $^{5}$Huawei Heisenberg Research Center (Munich)
}

\begin{document}

\makeatletter
\let\@oldmaketitle\@maketitle% Store \@maketitle
\renewcommand{\@maketitle}{\@oldmaketitle% Update \@maketitle to insert...
  \begin{center}
    {\includegraphics[width=1\linewidth]{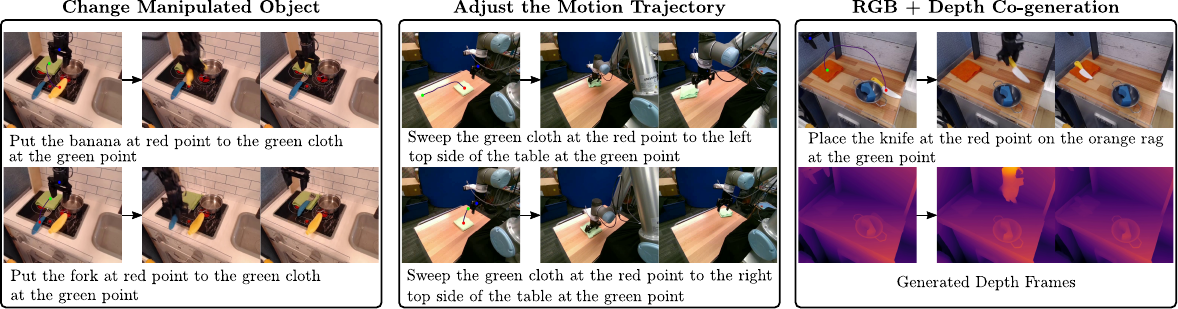}\par
    \captionsetup{type=figure}
     \captionof{figure}{ Capabilities of our proposed model, \modelname, which supports a variety of features, including changing the manipulated object, adjusting motion trajectories, and co-generating RGB and depth videos. }
     \label{fig:teaser}}
  \end{center}
  \vspace{-1.5em}
    }
\makeatother

\maketitle
\addtocounter{figure}{-1}

\thispagestyle{empty}
\pagestyle{empty}
%%%%%%%%%%%%%%%%%%%%%%%%%%%%%%%%%%%%%%%%%%%%%%%%%%%%%%%%%%%%%%%%%%%%%%%%%%%%%%%%

\begin{abstract}
Video diffusion models provide powerful real-world simulators for embodied AI but remain limited in controllability for robotic manipulation. Recent works on trajectory-conditioned video generation address this gap but often rely on 2D trajectories or single modality conditioning, which restricts their ability to produce controllable and consistent robotic demonstrations. We present \modelname, a depth-aware trajectory-conditioned video generation framework that extracts multiple orthogonal representations from the input trajectory, capturing depth, semantics, shape and motion, and injects them into the diffusion model. Moreover, we propose to jointly generate spatially aligned RGB and depth videos, leveraging cross-modality attention mechanisms and depth supervision to enhance the spatio-temporal consistency. Finally, we introduce a multimodal policy model conditioned on the generated RGB and depth sequences to regress the robot's joint angles. Experiments on Bridge V2, Berkeley Autolab, and simulation benchmarks show that \modelname $ $ achieves superior visual fidelity and consistency while yielding higher manipulation success rates compared to existing baselines.
\end{abstract}
%%%%%%%%%%%%%%%%%%%%%%%%%%%%%%%%%%%%%%%%%%%%%%%%%%%%%%%%%%%%%%%%%%%%%%%%%%%%%%%%

\input{introduction}

\input{related_work_new}

\input{method_new}
\input{Experiments}

\section{CONCLUSIONS}
We propose \modelname, a depth-aware trajectory-conditioned video generation framework for robotic manipulation. Our approach advances three fronts: (1) depth-aware 3D trajectories enable precise control of both robot arms and manipulated objects; (2) object-centric DINOv2 features enhance geometric and motion consistency, avoiding trajectory drift, object disappearance, and distortion observed in baselines; (3) RGB and depth videos co-generation improves visual quality and provides multimodal data that boosts downstream policy performance, as confirmed by ablation studies. Together, these contributions yield more accurate, coherent, and controllable robotic manipulation videos, validated across open-source benchmarks and simulator datasets.

% TODO Write Limitations not Future Works
While promising, challenges remain. Our model does not support long-horizon  controllability. Moreover, we currently support the controllability of one object only in the scene. Future works can address generation conditioned on multiple trajectories in parallel or sequentially to enable multi-object manipulation.

% \clearpage

{
\bibliographystyle{IEEEtran}
\bibliography{references}
}

%%%%%%%%%%%%%%%%%%%%%%%%%%%%%%%%%%%%%%%%%%%%%%%%%%%%%%%%%%%%%%%%%%%%%%%%%%%%%%%%

%%%%%%%%%%%%%%%%%%%%%%%%%%%%%%%%%%%%%%%%%%%%%%%%%%%%%%%%%%%%%%%%%%%%%%%%%%%%%%%%



% \section*{ACKNOWLEDGMENT}

% The preferred spelling of the word ÒacknowledgmentÓ in America is without an ÒeÓ after the ÒgÓ. Avoid the stilted expression, ÒOne of us (R. B. G.) thanks . . .Ó  Instead, try ÒR. B. G. thanksÓ. Put sponsor acknowledgments in the unnumbered footnote on the first page.

% %%%%%%%%%%%%%%%%%%%%%%%%%%%%%%%%%%%%%%%%%%%%%%%%%%%%%%%%%%%%%%%%%%%%%%%%%%%%%%%%

% References are important to the reader; therefore, each citation must be complete and correct. If at all possible, references should be commonly available publications.

\end{document}

%% file: introduction.tex
\section{INTRODUCTION}
The growing interest in embodied intelligence has catalyzed extensive research on video-based learning methods for robotic arm training~\cite{learning2imitate,learningbywatching,Learningmulti-stage,sharma19thirdperson,schmeckpeper2020rlv,skill,roboenvision}. Recent frameworks such as Vid2Robot~\cite{vid2robot} have shown that policies can be learned directly from videos. However, collecting large-scale datasets with real robots is costly and often requires teleoperation or human supervision to ensure safe manipulation. Recently, video diffusion models~\cite{du2023learning} have  shown an excellent ability to synthesize robotic demonstrations, thereby enriching training data and enabling the development of more capable video-based policy models~\cite{chi2023diffusionpolicy,act} and/or Vision-Language-Action models (VLAs)~\cite{openvla,black2024pi_0}.

Conventional video generation models typically rely on natural language prompts as conditions~\cite{opensora2,wan}. While effective for general-purpose video synthesis, language alone often lacks the precision needed to describe detailed robotic motions or manipulation trajectories. In robotic video generation, fine-grained controllability is crucial to ensure alignment with physical constraints and task requirements. Therefore, multimodal inputs such as sketches, object start/end points, or explicit motion trajectories are needed to guide the generation process and achieve more accurate and controllable video synthesis.

Recent works such as Tora~\cite{tora}, DragAnything~\cite{draganything}, and FreeTraj~\cite{freetraj} explore 2D trajectory-conditioned video generation by either introducing motion encoders, conditioning on the VAE latents of displaced objects, or modifying the initial noise sampling. However, robotic arm motion is inherently defined in 3D space, making purely 2D guidance insufficient. Incorporating depth information can improve controllability and better handle occlusions and disocclusions in image space. Methods like LeviTor~\cite{levitor} take a step toward integrating 3D trajectory information, but when applied to robotic demonstrations, they often lack explicit action control, resulting in inconsistent interactions between the arm and objects. Achieving precise 3D trajectory control in the generated video while preserving object and interaction consistency thus remains an open challenge for current video generation models. It requires a more holistic approach that captures all relevant aspects of the trajectory.

\textbf{Contributions:} To address these limitations, we propose \modelname, a novel framework that encodes multiple complementary trajectory representations inside a video diffusion model, enhancing its trajectory-following capabilities. Instead of relying one one trajectory representation like previous works, we fuse depth-encoded 3D trajectories, high-level semantic DINOv2 object features and pixel-coordinates augmented instruction prompts. Moreover, our model co-generates RGB and depth videos. Our insight is that  adding multimodal generation capabilities to the model improves the video consistency through depth supervision. Finally, we propose a multimodal policy model conditioned on the generated video and depth sequence, enabling the regression of joint control commands of robotic arms. In summary, our contributions are: 
\begin{enumerate}
\item We propose a trajectory-conditioned video diffusion model that integrates complementary trajectory representations into the diffusion transformer (DiT), namely: depth-aware 3D trajectories, high-level semantic features, and  coordinate augmented text captions. 

\item We introduce an object-centric feature control method, which extracts DINOv2 features of the manipulated object from the first frame and propagates them along the corresponding trajectory in subsequent frames. This encoding helps capture shape, semantics, and motion information, and is further injected inside the DiT.
%enhancing geometric consistency and realism in object–robot interactions.
\item We co-generate depth sequences alongside RGB videos, using depth supervision to enforce constraints and improve the quality of the generated videos. Moreover, we propose a multimodal policy model (RGB + Depth videos) for robotic manipulation, enabling more consistent learning.
\item Our experiments show that our framework produces better video quality and higher task success rates than previous trajectory-conditioned video methods, demonstrating its effectiveness for real-world robotic learning.
\end{enumerate}

%% file: related_work_new.tex
\section{RELATED WORK}

\subsection{Video Diffusion for Robotics}
% Diffusion-based generative models have greatly advanced image and video synthesis. Early methods such as DDPM~\cite{ho2020denoising} achieved high-quality image generation, while latent diffusion models (LDMs)~\cite{stablediffusion} improved efficiency by operating in a compressed latent space. Building on these foundations, text-to-video (T2V) frameworks~\cite{blattmann2023align, ho2022imagen} and image-to-video (I2V) methods such as VideoCrafter2~\cite{chen2024videocrafter2}, CogVideoX~\cite{yang2024cogvideox}, and SVD~\cite{blattmann2023stable} enable the synthesis of dynamic scenes from textual descriptions or static images.

Video generation~\cite{blattmann2023align, ho2022imagen,chen2024videocrafter2, yang2024cogvideox,blattmann2023stable} has recently emerged as a promising tool for robotic manipulation, serving as a world model, simulator, and data augmentation source. Early studies used generated videos to predict future robot actions and states, enabling policy learning~\cite{learning2imitate,learningbywatching}. Other works incorporate gestures, text instructions, or robot end-effector’s 3D trajectory-based control~\cite{irasim}, which relies on precise motion data that is often difficult to obtain in practice. Large-scale synthetic datasets such as UniPi~\cite{du2023learning} and UniSim~\cite{yang2023learning} have been proposed to support scalable robot learning, while model-based approaches like DreamerV3~\cite{hafner2023mastering} and DayDreamer~\cite{wu2023daydreamer} further combine latent world models with inverse dynamics to predict rewards and generate optimal actions. Despite these advances, many existing methods still face challenges in achieving fine-grained controllability.

In contrast, our method uses object-level trajectories, which are easier to acquire and allow explicit control over both robot arms and objects, providing more precise and interpretable manipulation videos.

\subsection{Trajectory Control in Video Generation}
Recent advances in controllable video generation have sought to improve motion precision while retaining the generative power of diffusion models. Early approaches employed optical flow or recurrent networks to encode object trajectories~\cite{tora,draganything}, but achieved only coarse control. Subsequent methods introduced bounding-box representations (e.g., TrailBlazer~\cite{trailblazers}) and trajectory vector maps (e.g., MotionCtrl~\cite{motionctrl}, DragNUWA~\cite{dragnuwa}) for instance-level guidance, while entity-aware methods such as DragAnything exploited segmentation masks for higher-fidelity motion. To address the limitations of 2D control, approaches like 3DTrajMaster~\cite{trajmaster} and LeviTor~\cite{levitor} incorporated depth cues and 6-DoF pose sequences to enable more realistic object manipulation. Multi-object trajectory control has also been explored for better coordination in complex scenes.

Both training-based and training-free paradigms have been investigated. Methods such as Tune-A-Video~\cite{tune} and MotionDirector~\cite{motiondirector} leverage reference videos to generalize motion patterns, whereas training-free approaches like Peekaboo~\cite{peekaboo}, TrailBlazer~\cite{trailblazers}, and DragAnything~\cite{draganything} use attention modulation or explicit trajectory guidance during inference, reducing annotation costs. Yet both methods face challenges when applied to robot demonstrations, particularly in maintaining consistency during robot-object interactions.

Building on these insights, our work trains a robotic arm video generation model with several explicit and implicit trajectories representations (depth, high-level semantics, corrdiantes) to achieve fine-grained controllable motion. 

%This design improves geometric and motion consistency of manipulated objects, strengthens interaction fidelity, and enhances temporal coherence, yielding flexible and high-quality video generation suitable for simulation, policy learning, and data augmentation.

%% file: method_new.tex
\section{METHOD}
\input{Figures/methods}
\label{sec:method}
This section introduces the key components and functionality of \modelname. As shown in Fig. \ref{fig:method}, \modelname $ $ supports two output modalities, RGB and depth videos and incorporates three forms of control signals: depth-awared trajectories, object-level DINOv2 features and text prompt augmented with pixel coordinates. We first review our Task Formulation (Sec. \ref{subsec:task}), followed by the details of our Representation Encoding pipeline (Sec. \ref{subsec:data}) our Control Injection Module (Sec. \ref{subsec:injection module}), and Multimodal Generation (Sec. \ref{subsec:multimodal generation}) and Multimodal Policy Model (Sec. \ref{subsec:policy model}).

\subsection{Task Formulation and Model Overview}
\label{subsec:task}
\noindent\textbf{Task Formulation:} Our goal is to synthesize a realistic manipulation video with $\textit{N}$ frames, $\textit{V}\in \mathbb{R}^{N \times 3 \times H \times W}$, given an initial frame $\textit{I}\in \mathbb{R}^{3 \times H \times W}$ containing task-relevant objects, a binary mask $\textit{M}\in \mathbb{R}^{H \times W}$ specifying the manipulated object, a user-defined textual prompt $\textit{c}$, and an object trajectory sequence $\textit{q}=\{(x_i,y_i,d_i)\}_{i=0}^{N-1}$, where $\textit{x}$ and $\textit{y}$ are pixel coordinates relative to the image origin and $\textit{d}$ represents the relative depth value at each frame $\textit{i}$. The trajectory sequence $\textit{q}$ explicitly defines the manipulated object's motion from a starting point (e.g. for picking) to a target location (e.g. for placing), providing direct control over the task.

\noindent\textbf{Model Overview:} We design our model as a conditional multimodal Latent Diffusion Model (LDM). We encode different representations of the input trajectory, capturing different but complementary aspects of the video to be generated: depth, motion, object semantics and shape. Namely, we extract three representations: $\mathcal{D}=\{z_0^{\text{ref}}, y_{dino}, y_c\}$, where:
\begin{itemize}
    \item $z_0^{\text{ref}}$ is the latent representation of the first video frame with the color-coded depth-aware trajectory.
    \item $y_{dino}$ represents the sequence of DINOv2 features of the manipulated object, propagated along its trajectory.
    \item $y_c$ is the text prompt augmented with detailed pixel-level positional information derived from the trajectory.
\end{itemize}

To encode the videos into a latent space, we employ a pre-trained open-source 3D causal variational autoencoder (VAE) $\mathcal{E}$. For a video  $V = \{I_i\}_{i=0}^{N-1}$ of dimension $\mathbb{R}^{3 \times N \times H \times W}$, where $I_i$ denotes $i^{th}$ image frame, $N$ represents the number of frames, and $H$ and $W$ are the height and width of each frames, the VAE outputs a latent feature $z$ of dimension $\mathbb{R}^{16 \times n \times h \times w}$, where $h$ and $w$ refer to the height and width in the latent space and $n$ represents the number of frames after the VAE’s temporal compression. 

% To reconstruct the original video $V$ during training, we first process it by the VAE to obtain $z$ and apply the forward diffusion process by progressively adding Gaussian noise to $z$ over $T$ steps, generating a sequence of noisy samples $\{z_{0}, z_{1}, \dots, z_{f-1}\}$.

We thus train the denoising network $\epsilon_{\theta}(\cdot, t)$, implemented as a DiT, to learn to progressively remove the added noise and reconstruct the video sequence given the set of input conditions. The training objective is:
\[
\mathcal{L}_{\text{diffusion}} = \mathbb{E}_{z, \mathcal{D}, \epsilon \sim \mathcal{N}(0,1), t} \left\| \epsilon - \epsilon_{\theta}(z_t, \mathcal{D}, t) \right\|_2^2
\]

\subsection{Encoding Trajectory Representations}  
% \mo{It would be better if these subsection refer to modules in the main diagram image. so maybe in the main picture, currently Fig 1, you make labels of the modules, with surrounding squares, or background colors, ex: control injection module ...} 
\label{subsec:data}

% We describe how we extract the conditioning signals from the groundtruth video that are injected inside the DiT training.

To obtain the trajectory of the manipulated object, we first identify the object name from the caption and extract its binary mask in each frame using Grounded-SAM\cite{ren2024grounded} and TrackAnything \cite{yang2023track} (see Fig. \ref{fig:data_process}).  
Given the sequence of object masks $M = \{M_i\}_{i=0}^{N-1}$, we compute the center of each mask to derive pixel-relative 2D coordinates $\{(x_i,y_i)\}_{i=0}^{N-1}$ forming the 2D trajectory sequence. Next, we employ the depth estimation network Video Depth Anything \cite{video_depth_anything} to predict relative depth video $V_{\text{depth}} = \{I_i^{depth}\}_{i=0}^{N-1}, \quad I_i^{depth} \in \mathbb{R}^{3 \times H \times W}$, which matches the resolution and frame count of the original RGB video. Our approach does not require highly accurate metric depth measurements, simplifying user interaction during inference. Finally, we associate each 2D trajectory point with its corresponding depth value \(d_i\) extracted from the corresponding depth video:
\[
d_i = I_i^{depth}(x_i, y_i).
\]
By combining the 2D coordinates with estimated depth values, we construct depth-aware control trajectory sequences.
\[
q = \{(x_i, y_i, d_i)\}_{i=0}^{N-1}.
\]
\noindent\textbf{Text Prompt.} We provide a detailed description of the robotic arm task in the form of a textual prompt (e.g., ``The robotic arm at blue point $(x_{\textrm{robot}},y_{\textrm{robot}})$ moves to the object at red point $(x_{\textrm{obj}},y_{\textrm{obj}})$, picks it up, and then moves to green point $(x_{\textrm{des}},y_{\textrm{des}})$''), where all point coordinates are pixel-relative. The text prompt is encoded using the T5 model~\cite{t5}, producing a feature vector $y_c$ that is incorporated into the denoising network $\epsilon_{\theta}$ through cross-attention. The caption provides a representation for the trajectory on a coarse level.

\input{Figures/data_process}

\noindent\textbf{Reference Image.} We generate a reference image $I_0^{\textrm{ref}}$ in the following way: we draw the  object’s start and end points as well as the trajectory line color-coded with the depth value on the first frame $I_0$. $I_0^{\textrm{ref}}$ is then encoded by a VAE to obtain latent feature representations $z_0^{\textrm{ref}} \in \mathbb{R}^{16 \times 1 \times h \times w}$. The latent feature $z_0^{\textrm{ref}}$ is then concatenated time-wise with $z$, and then fed into the denoising network $\epsilon_\theta$. This reference-frame latent feature encodes depth and motion providing a conditioning signal that propagates to other frames through self-attention.

\noindent\textbf{DINOv2 Object Features.} To provide object-centric semantic and motion information, we use a sequence of DINOv2 features as a key condition. First, we extract the manipulated object region from the initial frame $I_0$ using its mask $M_0$. To maximize the retention of object features and minimize background influence, we crop this region using the bounding box size and extract DINOv2 features from it. The DINOv2 model outputs features from its last layer that are rich in high-dimensional semantic information, which is essential for accurate object modeling.

We process the extracted object DINOv2 features on a blank background and resize them to match the latent space dimensions. This preserves object information while suppressing background noise. We then propagate these features along the object’s motion trajectory $q$, placing them at their corresponding locations in each frame to create a feature map of the same size as the video latent. This representation encodes the object’s semantics, shape, and motion.

Finally, we apply temporal interpolation to the DINOv2 features, compressing the time dimension to match that of the latent representation. The final feature has the same temporal and spatial dimensions as the latent space after the VAE process $y_{dino} \in \mathbb{R}^{ 1024 \times n \times h \times w}.$

\subsection{Control Injection Module}
\label{subsec:injection module}
We design a DINOv2 patch embedding module that first performs spatio-temporal downsampling on the input $y_{dino}$ to match the dimensions of the latents features after patch embedding $y_{dino} \in \mathbb{R}^{C \times (n/2 \times h/2 \times w/2)}$. 

Then, we inject the patchified DINOv2 features into the DiT backbone using a specialized Fusion Block. This module employs a gating mechanism to selectively modulate the DINOv2 features, followed by LayerNorm to stabilize the output. The modulated features are then residually integrated with the Transformer's hidden states. This allows for the progressive fusion of the DINOv2 features into each DiT block while preserving the integrity of the original hidden state. The fusion process is defined by:
\[
\quad h' = h + \text{LayerNorm}(y_{dino} \odot G) \odot (y_{dino} \odot G)
\]
\[
\quad G = \sigma(W_g y_{dino} + b_g)
\]
where $h$ denotes the Transformer hidden states, $\sigma$ is the Sigmoid activation function, $\odot$ denotes element-wise multiplication, $W_g$ and $b_g$ are the learnable weights and biases of the gating mechanism, $G$ is the gating mask, and $h'$ is the fused output. 

%after gating, normalization, and residual integration.

\subsection{Multimodal Generation}
\label{subsec:multimodal generation}
We incorporate the depth video, generated by Video Depth Anything, as an additional modality to leverage its rich spatial information (as illustrated in Fig.~\ref{fig:method}). By simultaneously generating both depth and RGB frames, the model is able to produce robust spatial information while preserving the scene's contextual structure. During training, the model applies self-attention across both modalities, enabling it to jointly capture and leverage the complementary information from the RGB and depth video streams.

To encode the depth information, we use the same VAE we use for the RGB frames. The two modalities are concatenated along the temporal dimension at the input, rather than along the channel dimension. In this setting, the input is effectively a single long sequence formed by concatenating the RGB and depth videos in time. This approach has the advantage of avoiding the need for additional embedding layers and separate convolution projection layers to predict independent depth noise, which reduces training complexity. Consequently, the final denoiser is formulated as:
\[
D_\theta\big((z \Vert \, z_{\textrm{depth}}); \sigma, \{z_0^{ref}, y_{dino}, y_{c}\}\big).
\]

\input{Tables/comparison}

\input{Tables/comparison2}
\subsection{Multimodal Policy Model}
\label{subsec:policy model}
We propose an architecture see right part of Fig. ~\ref{fig:data_process} to estimate robot joint states from RGB and depth videos. RGB and depth videos are first encoded using VAE to obtain latent representations, which are then processed separately through their respective patch embedding. Each latent feature is first processed through a spatial transformer and then a temporal transformer to capture spatial and temporal dependencies. The two features then interact through a cross-attention block, exchanging information, and are finally summed before being decoded through a ResNet-based decoder to predict the robot’s gripper state and joint angles.

%% file: Figures/methods.tex
\begin{figure*}[ht]
  \centering
  \vspace{1em}
  
  \includegraphics[width=0.96\textwidth]{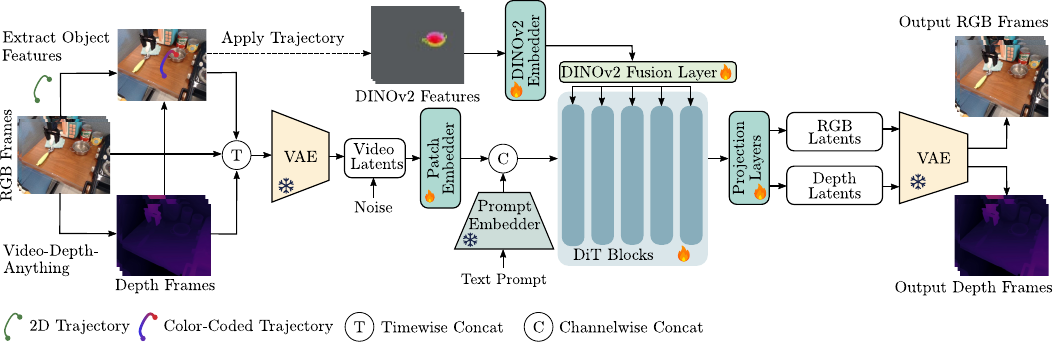}
  \caption{Model Architecture. We extract several representations from the trajectory (DINOv2 object-level features, pixel coordinates augmented prompt and reference image with the depth-aware trajectory overlaid. DINOv2 features are injected into each block through DINOv2 fusion Layers. To enable multimodal output generation, we concatenate the reference frame, RGB frames, and depth frames in the time dimension, and process them with the same patch embedding module.}
  \label{fig:method}
  \vspace{-2em}
\end{figure*}

% \caption{

% \fengyi{add a name or a global description to the figure here. }

% Our model presents a framework that takes as input a reference frame along with noisy latent representations of both RGB and depth modalities. The denoising network incorporates multiple conditioning signals ($D$), where DINOv2 features are injected into each block through a Dino fusion Layer. To support multimodal output generation, we time-wise concatenate the reference frame, RGB frames, and depth frames, enabling a shared one patch embedding module to process the combined inputs. The final predictions are produced through convolution-based projection layers.  

% \mo{Is this the latest figure? because the paper/contributions talk about 3D trajectory control, while the image does not mention/show any relevance to trajectory control . may be you need to add it explicitly instead of letting the readers/reviewers infer that it comes form the depth / dino. } 

% \fengyi{exactly, this figure should be modified in a way that it highlights your two contributions in intro at first glance, while still showing the whole framework.} 

% \george{yes that's important if you add depth into the trajectory it should be shown how} }

%% file: Figures/data_process.tex
\begin{figure*}[ht]
  \centering
  \vspace{1.05em}

  \includegraphics[width=\textwidth]{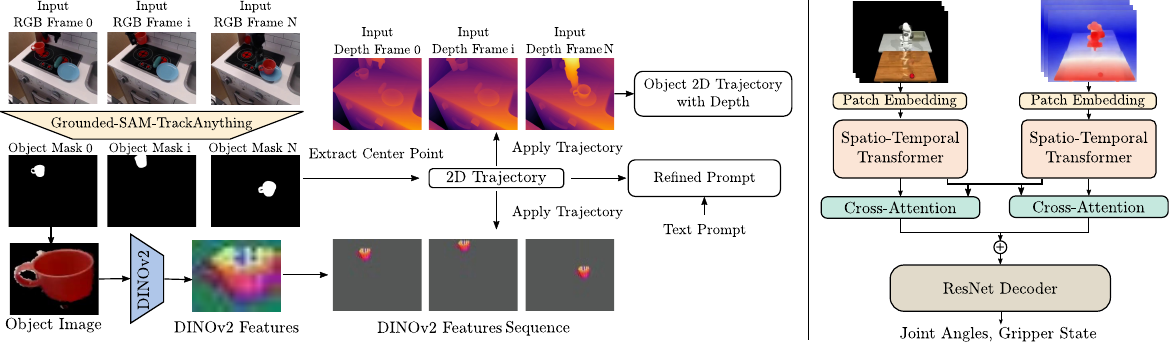}
  \caption{Left: Representation Encoding. We first extract object masks $M$ using Grounded-SAM-TrackAnything to obtain the 2D trajectory of the manipulated object and augment it with relative depth estimates. In parallel, we extract DINOv2 features from the object and align them to the trajectory coordinates. We augment the text prompt with the start and end positions. Right: utilize modal Policy Model to predict robot joint angles and gripper state}
  \label{fig:data_process}
  \vspace{-2em}
\end{figure*}

% \caption{
% \fengyi{add a name or a global description to the figure here. }

% Given a reference image $I$ and a text prompt, our framework generates a robotic manipulation video $V$ guided by a designed 2D trajectory enriched with depth information. Specifically, we first extract object masks 
% $M$ using Grounded-SAM-TrackAnything to obtain the 2D trajectory of the manipulated object. We then enhance this trajectory with relative depth estimates obtained from Video-depth-Anything, resulting in a depth-aware 2D trajectory. In parallel, we extract DINOv2 features from the object image, capturing high-dimensional representations of its appearance and shape. These features are resized and aligned to the latent coordinates according to the computed trajectory, producing a sequence of DINOv2 feature maps. 

% This sequence is injected into the DiT blocks, providing rich appearance and shape cues to enhance identity consistency and visual fidelity throughout the generated video. 

% \mo{the meaning of Apply trajectory is not clear. The second part of the image can be enhanced with better flow of arrows} 

% \george{add the color encoded trajectory in the teaser figure and in this figure}  }

%% file: Tables/comparison.tex
\begin{table*}[!ht]
    \centering
    \vspace{1em}
    \setlength{\tabcolsep}{3pt}
    \renewcommand\arraystretch{1}
    \adjustbox{width=0.8\linewidth}{%
    \begin{tabular}{l|l|cccc|c|c}
    \hline
        \multirow{2}{*}{Dataset} & \multirow{2}{*}{Method} & \multicolumn{4}{c|}{Vbench Evaluation} & \multicolumn{1}{c|}{Trajectory Deviation} & \multicolumn{1}{c}{Task Evaluation} \\
        ~ & ~ & Mot.Cons.$\uparrow$ & Bg.Cons.$\uparrow$ & Subj.Cons.$\uparrow$ & Tem.Fli.$\uparrow$ & Object Traj. Error $\downarrow$ & Success Rate(\%)$\uparrow$ \\
    \hline
        \multirow{5}{*}{Bridge V2} 
            & LeviTor      & 0.9712 & 0.9289 & 0.9272 & 0.9817 & 46.52 & N/A \\
            & Tora         & 0.9875 & 0.9507 & 0.9346 & 0.9821 & 35.67 & N/A \\
            & MotionCtrl   & 0.9792 & 0.9471 & 0.9317 & 0.9811 & 38.24 & N/A \\
            & DragAnything & 0.9810 & 0.9442 & 0.9289 & 0.9832 & 37.11 & N/A\\
            & \textbf{Ours} & \textbf{0.9891} & \textbf{0.9512} & \textbf{0.9383} & \textbf{0.9849} & \textbf{25.30} & N/A\\
    \hline
        \multirow{5}{*}{Berkeley UR5} 
            & LeviTor      & 0.9803 & 0.9436 & 0.9325 & 0.9761 & 47.29 & N/A\\
            & Tora         & 0.9818 & 0.9502 & 0.9410 & \textbf{0.9833} & 35.73 & N/A \\
            & MotionCtrl   & 0.9844 & 0.9472 & 0.9391 & 0.9761 & 37.91 & N/A\\
            & DragAnything & 0.9827 & 0.9488 & 0.9402 & 0.9761 & 39.76 & N/A\\
            & \textbf{Ours} & \textbf{0.9845} & \textbf{0.9509} & \textbf{0.9417} & \textbf{0.9833} &\textbf{22.37} & N/A\\
    \hline
        \multirow{5}{*}{Simulator} 
            & LeviTor      & 0.9711 & 0.9394 & 0.9381 & 0.9819 & 33.52 & 0.0\\
            & Tora         & 0.9844 & 0.9441 & 0.9452 & 0.9803 & 35.44 & 36.8\\
            & MotionCtrl   & 0.9832 & 0.9387 & 0.9392 & 0.9813 & 29.83 & 29.6\\
            & DragAnything & 0.9811 & 0.9424 & 0.9433 & 0.9811 & 30.27 & 31.2\\
            & \textbf{Ours} & \textbf{0.9865} & \textbf{0.9473} & \textbf{0.9495} & \textbf{0.9821} & \textbf{19.88} &\textbf{65.2} \\
    \hline
    \end{tabular}}
    \caption{Main comparison of trajectory-conditioned video generation models. Bold numbers indicate the best performance.}
    \label{tab:baseline_comparison}
    \vspace{-2em}
\end{table*}

%% file: Tables/comparison2.tex
\begin{table*}[!ht]
    \centering
    \vspace{1em}
    \setlength{\tabcolsep}{3pt}
    \renewcommand\arraystretch{1}
    \adjustbox{width=\linewidth}{%
    \begin{tabular}{l|l|cccc|cccc|c}
    \hline
        \multirow{2}{*}{Dataset} & \multirow{2}{*}{Method} & \multicolumn{4}{c|}{Vbench Evaluation} &\multicolumn{4}{c|}{Depth Deviation} & \multicolumn{1}{c}{Trajectory Deviation}\\
        ~ & ~ & Mot.Smth..$\uparrow$ & Bg.Cons.$\uparrow$ & Subj.Cons.$\uparrow$ & Tem.Fli.$\uparrow$  & LPIPS $\downarrow$ & SSIM $\uparrow$ & PSNR $\uparrow$ & FVD$\downarrow$ & Object Traj. Error $\downarrow$\\
    \hline
        \multirow{6}{*}{Bridge V2} 
            & first frame (RGB)      & 0.9777 & 0.9183 & 0.9155 & 0.9817 & N/A & N/A & N/A & N/A & 39.88\\
            & first frame (RGB and Depth)      & 0.9813 & 0.9202 & 0.9164 & 0.9821 & 0.3480 & 0.7022 & 15.33 & 222.7 & 45.27\\
            & point frame         & 0.9892 & 0.9221 & 0.9268 & \textbf{0.9849} & 0.3449 & 0.6822 & 15.12 & 196.6 & 36.41\\
            & 2D traj. frame   & 0.9864 & 0.9369 & 0.9373 & 0.9804 & 0.3485 & 0.6643 & 14.98 & 181.8 & 29.67\\
            & 3D(depth) traj. frame & 0.9887 & 0.9509 & 0.9377 & 0.9845 & 0.3199 & 0.7305 & 15.86 & \textbf{169.4} & 26.81\\
            & \textbf{3D(depth) traj.\&Dino} & \textbf{0.9891} & \textbf{0.9512} & \textbf{0.9383} & \textbf{0.9849} & \textbf{0.3152} & \textbf{0.7326} & \textbf{15.91} & 177.1 & \textbf{25.30}\\
            
    \hline
            % & \textbf{CogvideoX-fun(3D(depth) traj.\&Dino)} & 0.9777 & 0.9119 & 0.8455 & 0.9761 & 0.9729 \\
            \multirow{6}{*}{Simulator} 
            & first frame (RGB)      & 0.9832 & 0.9018 & 0.9082 & 0.9818 & N/A & N/A & N/A & N/A & 61.32\\
            & first frame (RGB and Depth)      & 0.9854 & 0.9125 & 0.9091 & 0.9820 & 0.3972 & 0.4888 & 7.230 & 174.6 & 69.17\\
            & point frame         & 0.9855 & 0.9226 & 0.9130 & 0.9816 & 0.3959 &0.4906  & 7.331 & 163.05 & 26.83\\
            & 2D traj. frame   & 0.9852 & 0.9253 & 0.9317 & 0.9814 & 0.3889 & 0.4920 & 7.329 & 164.39 & 22.17\\
            & 3D(depth) traj. frame & 0.9851 & 0.9357 & 0.9396 & 0.9811 & 0.3889 & \textbf{0.4981}  & \textbf{7.350} & 159.45 & 21.39\\
            & \textbf{3D(depth) traj.\&Dino} & \textbf{0.9865} & \textbf{0.9473} & \textbf{0.9495} & \textbf{0.9821} & \textbf{0.3886} & \textbf{0.4981} & 7.347 & \textbf{158.44} & \textbf{19.88}\\
    \hline
    \end{tabular}}
    \caption{Ablation study on different different types of input condition.}
    \label{tab:ablation_comparison}
    \vspace{-2em}
\end{table*}

%% file: Experiments.tex
\section{EXPERIMENTS AND RESULTS}

To thoroughly evaluate the effectiveness of fine-grained control through trajectories and DINOv2 features, we conducted extensive experiments on multiple datasets and diffusion base model. Sec.~\ref{subsec:experiment setup} details the experimental setup, and Sec.~\ref{subsec:evaluation} presents the results and comparative analysis, then we do the ablation study in Sec.~\ref{subsec:ablation}
\subsection{Experimental setup}
\label{subsec:experiment setup}

\noindent\textbf{Datasets:}
We utilize four robotic arm datasets, including two publicly available datasets: BridgeDataV2(WidowX Robot)\cite{bridge} and Berkeley Autolab(UR5)\cite{BerkeleyUR5Website}. The third dataset was generated in the MuJoCo simulator with a Franka Panda robotic arm. In total, we curated approximately 50.6K videos. All datasets were preprocessed following the procedure outlined in the Method Sec.~\ref{sec:method}.

\noindent\textbf{Baseline Models:}
To evaluate the effectiveness of our method, we compare it against several state-of-the-art trajectory-controlled video generation approaches: LeviTor~\cite{levitor}, Tora~\cite{tora}, MotionCtrl~\cite{motionctrl}, DragAnything~\cite{draganything}. 

\noindent\textbf{Evaluation:}
We evaluate our approach on a total of 300 test samples, consisting of 100 samples from each dataset. The evaluation set covers diverse environments, robotic arms, manipulation instructions, and object types. Our evaluation is based on three criteria:  
\begin{enumerate}
    \item \textbf{Video Quality}: We use the following quantitative
    metrics for automatic evaluation: Motion Smoothness(Mot. Smth.), I2V Background consistency (Bg. Cons.), I2V Subject Consistency(Subj. Cons), Temporal flickering (Tem. Fli.). The metrics are calculated using VBench-2.0\cite{vbench} to assess the overall quality of the videos. 
    \item \textbf{Trajectory Accuracy}: We measure object trajectory fidelity by computing the Trajectory Error (Object Traj. Error), defined as the mean L1 distance between the input trajectories and those extracted from generated videos.  
    \item \textbf{Task Success Rate}: While not the focus of our work, we measure the efficiency of the multimodal policy model compared to the RGB only policy and the existing baselines. The RGB only policy model is used to evaluate task success rates for baselines on simulated datasets, as they do not generate depth videos. Furthermore, we compare task success rates when regressing trajectories from our generated videos using both RGB and depth versus only RGB.
    \item \textbf{Depth Video Quality}:We report LPIPS, SSIM, PSNR and FVD by comparing our generated depth videos with depth maps inferred from the generated RGB videos using Video Depth Anything.
    
\end{enumerate}

\noindent\textbf{Implementation details:} Our conditional video diffusion model is implemented based on the pre-trained CogVideoX-Fun-5B architecture~\cite{cogvideo}. We use a $69 \times 320 \times 320$ resolution video for both training and inference. The model is trained with the AdamW optimizer on 8 NVIDIA A800 GPUs for approximately 3 days, over a total of 50K training steps.

\input{Figures/Main_Comparisons_Qualitative}

\subsection{Qualitative and Quantitative Evaluation}
\label{subsec:evaluation}

\noindent\textbf{Quantitative evaluation:} 
Our model outperforms all baselines across all evaluation metrics, including video quality, trajectory accuracy, and task success rate on the simulation dataset(see Tab.~\ref{tab:baseline_comparison}). LeviTor, due to the lack of prompt-based control, fails to drive interactions between the robot arm and objects. Other methods can complete the tasks, but deformations occurring during object–robot interactions degrade video quality, resulting in low success rates.
% Our model outperforms all baselines across all evaluation metrics, including video quality, trajectory accuracy, and task success rate on the simulation dataset (see Table~\ref{tab:baseline_comparison}). LeviTor, due to the lack of prompt-based control, fails to drive interactions between the robot arm and objects. While other methods can complete the tasks, frequent deformations during object–robot interactions result in poor video quality, which leads to low success rates.

\noindent\textbf{Qualitative evaluation:} As shown in Fig.~\ref{fig:baseline_comparison}, our model produces the best visual quality among all methods. Competing approaches often struggle to move objects correctly along the intended trajectories and exhibit issues such as object disappearance, geometric distortions, or inconsistent appearances across frames. For example, in the first task of moving sushi, none of the baselines are able to follow the trajectory and move the object correctly. In the second task, while some models succeed in following the trajectory, the toy suffer from noticeable deformations. In the third task, MotionCtrl generates videos in which the bottle undergoes severe distortions. In the simulator task, TORA results in object disappearance, whereas DragAnything places the object in an incorrect location. In contrast, our method reliably follows the given trajectories while preserving object geometry, resulting in more coherent and realistic manipulation videos.

% As shown in Fig.~\ref{fig:baseline_comparison}, our model produces the best visual quality among all methods. Competing approaches often struggle to move objects correctly along the intended trajectories and exhibit issues such as object disappearance, geometric distortions, or inconsistent appearances across frames. In contrast, our method reliably follows the given trajectories while preserving object geometry, resulting in more coherent and realistic manipulation videos.
\input{Tables/success_rate_ablation}

\subsection{Ablation Study} 
\label{subsec:ablation}

To demonstrate the novel contributions and design choices of our method, we conduct ablation studies using different types of  input conditions to the DiT: First Frame RGB (without trajectory or point information, vanilla I2V), First Frame Multimodal (I2V with both RGB and depth videos), Point Image (object start and end points), 2D Trajectory Image (adding a 2D path on the point image), 3D Trajectory Image (adding depth information), and 3D Trajectory with DINOv2 features (combining depth trajectory and DINOv2 features). 
% Results are reported in Table~\ref{tab:ablation_comparison}, Fig.~\ref{fig:bridge_fig_ablations}, and Fig.~\ref{fig:ablations}.  
From Tab.~\ref{tab:ablation_comparison}, we can see that co-generating RGB and depth videos can help improve video quality. Moreover, using trajectory reference images further enhances video quality. Finally, depth-aware trajectories yield the best visual results and produce depth videos that are closer to those inferred by Video Depth Anything, highlighting the importance of depth information for fidelity.

We can see from Fig.~\ref{fig:bridge_fig_ablations},
inputs based on the first frame or points provide coarse control, making task completion difficult. For example, when executing the first task, conditioning on the first frame causes the robot arm to mistakenly manipulate the yellow cloth instead of the can, and conditioning on the point image also results in the object failing to move as expected. Similarly, As shown in Fig.~\ref{fig:ablations}, we observe that using the first frame does not move the bowl to the designated position, while using the point frame leads to severe deformation of both the robot arm and the object. As shown in Fig.~\ref{fig:bridge_fig_ablations}, while 2D and depth trajectories improve control precision, they still suffer from object inconsistency, such as appearance changes in the can and varying the shape of the bowl. In contrast, our final method—combining depth trajectories with DINOv2 features—not only follows the intended trajectories but also preserves object consistency, thereby achieving best video quality and trajectory accuracy. Task success rates are reported in Tab.~\ref{tab:success_rate_ablation}, showing that using RGB and depth videos for trajectory regression outperforms using RGB only. Furthermore, the 3D trajectory with DINOv2 features achieves the highest video quality, and the highest task success rate.

\input{Figures/bridge_fig_ablations}

\input{Figures/sim_fig_ablations}

%% file: Figures/Main_Comparisons_Qualitative.tex
\begin{figure*}[ht!]
  \centering
    \vspace{1.05em}
  \includegraphics[width=0.9\textwidth]{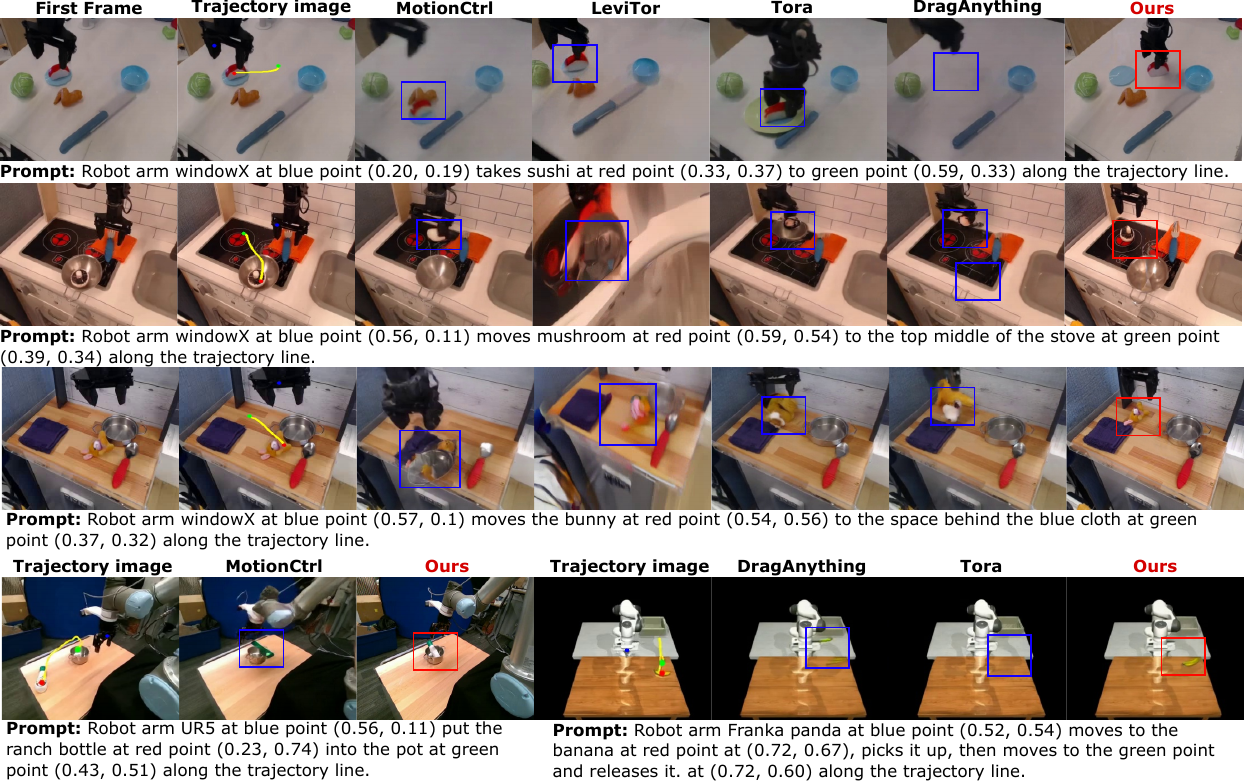}
  \caption{Qualitative comparison of trajectory-conditioned video generation approaches.}
  \label{fig:baseline_comparison}
  \vspace{-2em}
\end{figure*}

%% file: Tables/success_rate_ablation.tex
\begin{table}[!t]
    \centering
    \vspace{1em}
    \setlength{\tabcolsep}{3pt}
    \renewcommand\arraystretch{1}
    \adjustbox{width=\linewidth}{%
    \begin{tabular}{l|l|c}
    \hline
        \multirow{1}{*}{Policy Model} & \multirow{1}{*}{Condition} & Success Rate(\%)$\uparrow$ \\
    \hline
        \multirow{1}{*}{RGB} 
            & First Frame RGB       & 15.4 \\
    \hline
        \multirow{5}{*}{RGB \& Depth} 
            & First Frame Multimodal & 21.0 \\
            & Point Frame                & 35.6 \\
            & 2D Trajectory Image             & 52.4 \\
            & 3D  Trajectory Image      & 61.8 \\
            & \textbf{3D Trajectory w/ DINOv2} & \textbf{65.2} \\
    \hline
    \end{tabular}}
    \caption{Ablation study on different types of input condition's task success rate.}
    \label{tab:success_rate_ablation}
    \vspace{-2em}
\end{table}

%% file: Figures/bridge_fig_ablations.tex
\begin{figure*}[ht!]
  \centering
    \vspace{1.05em}
  \includegraphics[width=0.88\textwidth]{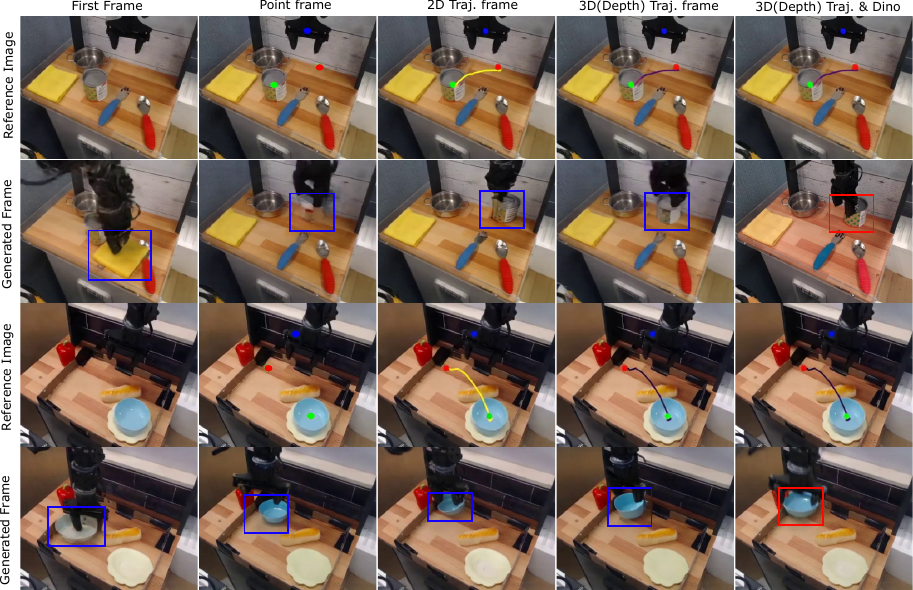}
  \caption{Qualitative Ablation study against different control approaches on the Bridge dataset}
  \label{fig:bridge_fig_ablations}
    \vspace{-1em}

\end{figure*}

%% file: Figures/sim_fig_ablations.tex
\begin{figure*}[ht!]
  \centering
  \vspace{1em}
  \includegraphics[width=0.9\textwidth]{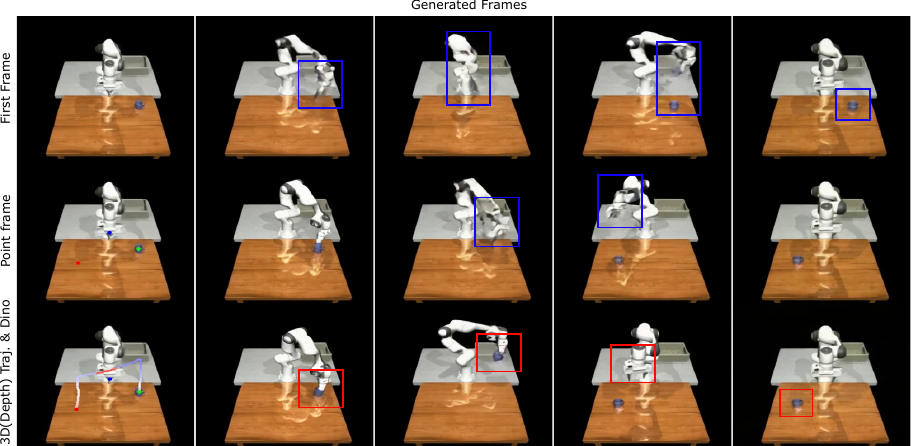}
  \caption{Qualitative Ablation study against different control approaches on the Simulation dataset}
  \label{fig:ablations}
  \vspace{-1em}
\end{figure*}